\documentclass[10pt,twocolumn,letterpaper]{article}

\usepackage[preprint]{iccv}      
\usepackage{blindtext}
\usepackage{dblfloatfix}
\definecolor{cvprblue}{rgb}{0.21,0.49,0.74}
\usepackage[pagebackref,breaklinks,colorlinks,allcolors=cvprblue]{hyperref}

\usepackage{tikz}
\usetikzlibrary{patterns}

\def\paperID{13836} 
\def\confName{ICCV}
\def\confYear{2025}

\newcommand\blfootnote[1]{%
  \begingroup
  \renewcommand\thefootnote{}\footnote{\noindent#1}%
  \addtocounter{footnote}{-1}%
  \endgroup
}

\usepackage{tcolorbox}
\title{How to Train your Text-to-Image Model:\\
Evaluating Design Choices for Synthetic Training Captions}

\author{ 
Manuel Brack$^{1,2}$\textsuperscript{\textdagger} \and Sudeep Katakol$^{1}$ \and Felix Friedrich$^{2,3}$\and Patrick Schramowski$^{2,3,4}$ \and Hareesh Ravi$^{1}$  \and Kristian Kersting$^{2,3,4}$ \and Ajinkya Kale$^{1}$ \and \vspace{-4mm}\\
\text{$\phantom{0}^{1}$Adobe Applied Research,$\phantom{0}^{2}$hessian.AI},$\phantom{0}^{3}$TU Darmstadt,$\phantom{0}^{4}$DFKI \\
{\tt\small mbrack@adobe.com}
}

\begin{document}
\maketitle
\begin{abstract}
Training data is at the core of any successful text-to-image models. The quality and descriptiveness of image text are crucial to a model's performance. Given the noisiness and inconsistency in web-scraped datasets, recent works shifted towards synthetic training captions. 
While this setup is generally believed to produce more capable models, current literature does not provide any insights into its design choices.

This study closes this gap by systematically investigating how different synthetic captioning strategies impact the downstream performance of text-to-image models. Our experiments demonstrate that dense, high-quality captions enhance text alignment but may introduce trade-offs in output aesthetics and diversity. Conversely, captions of randomized lengths yield balanced improvements across aesthetics and alignment without compromising sample diversity. 
We also demonstrate that varying caption distributions introduce significant shifts in the output bias of a trained model.
Our findings underscore the importance of caption design in achieving optimal model performance and provide practical insights for more effective training data strategies in text-to-image generation.
\end{abstract}
\vskip -5em

\section{Introduction}
\blfootnote{\textdagger Work partially done at DFKI \& TU Darmstadt}
Recently, large-scale, text-guided diffusion models (DM) have enabled versatile applications in image generation \cite{rombach2022High, ramesh2022hierarchical, saharia2022photorealistic, openai2023dalle3,liu2024playgroundv3improvingtexttoimage}. These models are trained to remove Gaussian noise from the images iteratively. At inference, we can generate novel images by initializing the reverse diffusion process from random noise. Combining these models with a versatile yet intuitive textual interface has been crucial to their success. These \textit{text-to-image} (T2I) models can generate images based on textual descriptions of a scene. Consequently, they are easily usable by expert and inexperienced users and can be employed for various tasks. Naturally, the performance of such models can mainly be expressed by the visual fidelity of the generated images and how well they correspond to the text prompt.

\begin{figure*}[t]
    \centering
\includegraphics[width=.75\linewidth]{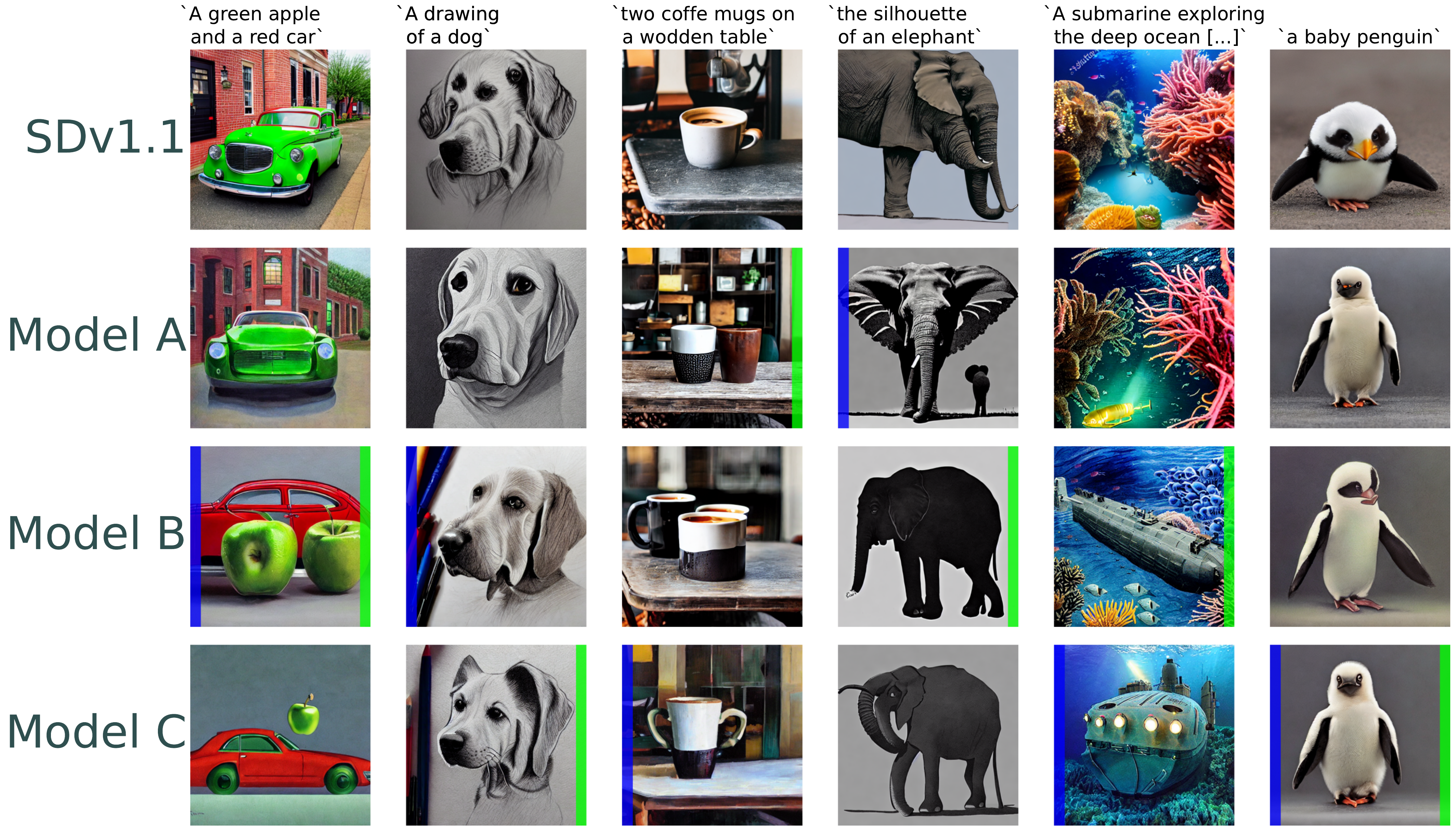}
\vskip -.25em
    \caption{Continual pre-training provides a suitable tool for investigating training design choices. We see clear improvements over the base model (SDv1.1) in image aesthetics and text following. Further, different caption distributions (Model A, B \& C) also yield measurable performance differences. Blue bars denote the highest aesthetics score over the 4 images whereas green shows the best pick-score.}
    \label{fig:continual_examples}
      \vskip -.75em
\end{figure*}
General-purpose T2I models require large-scale training datasets in the hundreds of millions or even billions of samples \cite{schuhmann2022laion,datacomp,schuhmann2021laion400m}. Given the task at hand, we require paired image-text data, where the textual captions should accurately describe associated images. 
Data at this scale is usually scraped from the internet, where text data can be collected from an image's ALT text in HTML. Arguably, the most prominent such dataset that is publicly available is LAION-5B \cite{schuhmann2022laion}. As the name suggests, LAION contains roughly 5B images, 2B of which come with corresponding English text captions. The LAION datasets or derivatives thereof have been used to train multiple T2I models \cite{rombach2022High,deepfloyd2023}. 

However, there are inherent limitations in the caption quality of web datasets. The ALT text descriptions from LAION are incredibly noisy and sometimes completely unrelated to the image content. Additionally, other data sources, like Common Catalogue \cite{gokaslan2023commoncanvas}, might provide high-quality images without any associated textual descriptions. 
Consequently, \citet{openai2023dalle3} proposed to use synthetic captions for T2I training. They argued that long, highly descriptive captions generated by a dedicated captioning model generally outperform shorter synthetic and original captions. However, we demonstrate that this general recommendation does not adequately cover relevant nuances and trade-offs in synthetic caption design.
Nonetheless, this paradigm has since become the de facto standard approach. For example, Stable Diffusion began incorporating synthetic captions for its third release \cite{esser2024scaling}, and Playgroundv3 \cite{liu2024playgroundv3improvingtexttoimage} was solely trained on synthetic captions. 

Despite the prevalence of this training paradigm, current literature offers little to no empirical information for key design choices. For example, some approaches develop dedicated captioning models \cite{liu2024playgroundv3improvingtexttoimage, openai2023dalle3}, while others report satisfactory results when using weak models \cite{gokaslan2023commoncanvas} like BLIP-2 \cite{li2023blip2}. Instead of following scientific rigor, practitioners often make these decisions based on intuition.    
However, a deeper scientific interpretation of these processes is instrumental in furthering our understanding of T2I models and improving current methods. 
This work bridges that gap by introducing a framework that provides valuable insights for T2I training with synthetic captions. Our structured investigation covers the most relevant aspects of synthetic caption design and derives novel recommendations and insights for practice.
Our results lead to the following observations that should inform the composition of T2I training data:

\begin{itemize}
    \item Using captions from stronger VLMs leads to better downstream text-following capabilities. 
    \item Highly descriptive, long captions are not unequivocally better than short, noisy ones. 
    \item For one, the former leads to \textit{bland} and non-diverse images for short text prompts at inference.
    \item For models with limited capacity, optimizing outputs aesthetic and T2I alignment are competing objectives. Consequently, training on dense captions results in less aesthetic outputs.
    \item This trade-off can be remedied by diversifying the length of captions used in training.
    \item We demonstrate that the distribution of key terms in training captions strongly influences output bias of the resulting model, for example, on gender.
\end{itemize}

\begin{figure*}[t]
\centering
    \includegraphics[width=.9\linewidth]{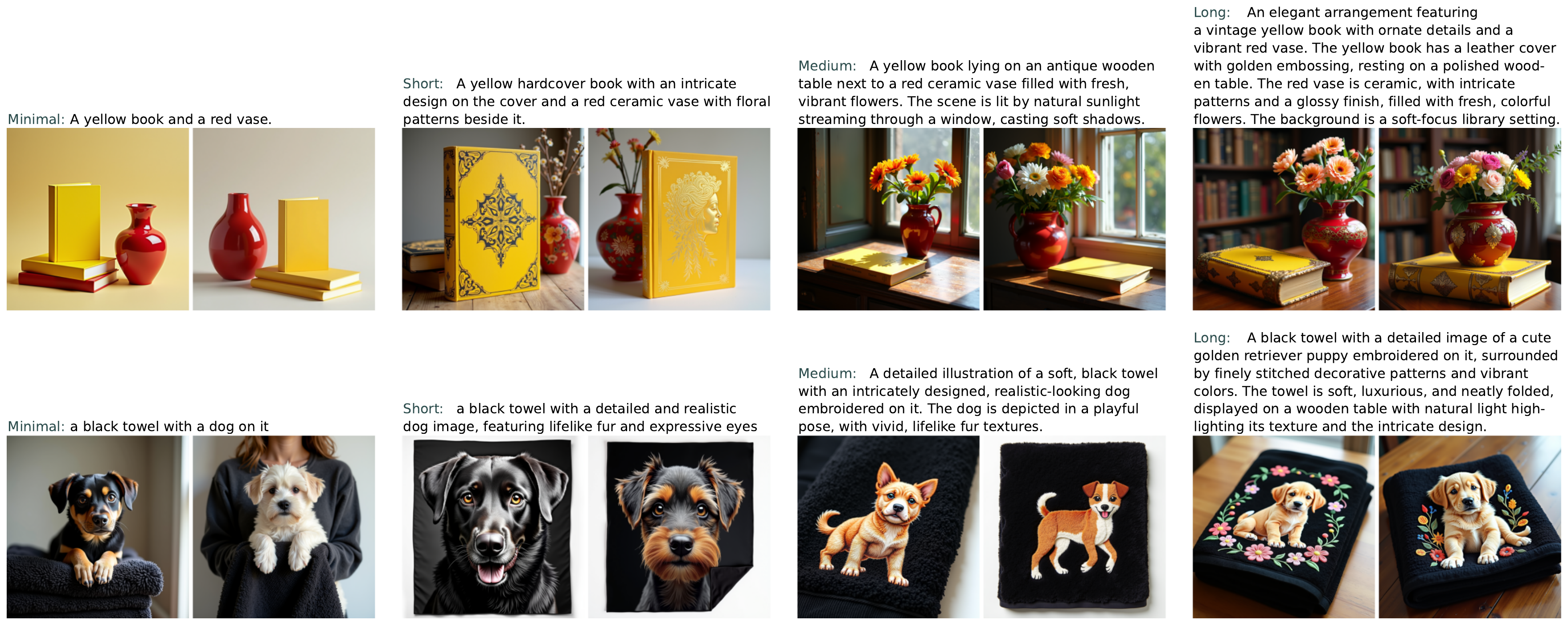}
    \vskip -.25em
    \caption{Qualitative examples of our evaluation prompt set. Each prompt is included in four increasingly detailed levels of instruction. All levels maintain the same subject. 
    Example images were generated with FLUX.1-dev.
}
    \label{fig:eval_prompts}
\end{figure*}

      \vskip -.75em
\section{Benefits and challenges of training on synthetic captions}

The advent of synthetic captioning in T2I model training has introduced a paradigm shift in the field, yet the scientific understanding of its nuances remains limited. To address this issue, we outline a set of hypotheses that systematically explore the impact of captioning strategies on model performance, followed by an experimental design that ensures rigorous empirical validation.

\subsection{Hypotheses \& Experimental Tests} \label{sec:hyp}
Building on recent advancements and informed by empirical gaps in the literature, we propose and systematically test the following hypotheses through rigorous experimental evaluation:

\noindent\textbf{\texttt{H1}: Long, dense captions yield better models.} 
\citet{openai2023dalle3} Prior work argued that training on highly descriptive captions generated by strong vision-language models (VLM) yields better T2I models.
Based on these suggestions, prior works have solely used synthetic captions from weak \cite{gokaslan2023commoncanvas} or strong VLMs \cite{liu2024playgroundv3improvingtexttoimage}. Conversely, other approaches retain some original captions and use synthetic ones for 50-90\% of the data \cite{sauer2024fast, openai2023dalle3}.

\textbf{\texttt{T1\!.\!1} Comparing noisy web-crawled captions and VLM captions.}
First, we evaluate the downstream performance of training on web-crawled captions and dense captions generated by increasingly more powerful VLMs.

\noindent\textbf{\texttt{H2}: Diversity in captions matter.} 
A potential issue with using a VLM to generate all training captions is a distinct lack of description diversity. Intuitively, this reduction to a narrow caption distribution may hurt the T2I models' generalization capabilities \cite{liu2024playgroundv3improvingtexttoimage, openai2023dalle3}. Therefore, we next evaluate multiple avenues for increasing caption diversity. 

\textbf{\texttt{T2\!.\!1} Diversity temperature sampling.}
As a naive baseline for this idea, we first increase the temperature of the VLM during caption sampling.

\textbf{\texttt{T2\!.\!2} Influence of caption density.} More recently, \citet{liu2024playgroundv3improvingtexttoimage} slightly diverged from suggestions of prior works and argued for training captions of varying density.
We compare different, narrow, and broad distributions of synthetic captions that only differ in density.

\textbf{\texttt{T2\!.\!3} Inter epoch diversity.} A more general variant of the prior setting further diversifies captions over different epochs. Thus, we never show the same image-caption pair during training. \citet{liu2024playgroundv3improvingtexttoimage} argue that this setup especially improves the T2I model when trained on limited amounts of data. 

\textbf{\texttt{T2\!.\!4} Explicit diversity through captioning personas.}
Instead of relying on different sampling strategies to achieve caption diversity, we evaluate a promising explicit setting. We prompt the VLM with crafted \textit{personas} with instructions on image aspects the respective captions should capture. For example, a \textit{Photographer} focuses on technical aspects like lighting, composition, focal points, exposure, and depth of field, while a \textit{Biologist} describes natural elements such as flora, fauna, ecosystems, or weather, with attention to scientific classifications and behaviors.
        
\noindent\textbf{\texttt{H3}: Gender distributions in captions influence T2I bias.} It is well-documented that generative models inherit stereotypical biases from their training data \cite{schramowski2022safe,caliskan2017semantics,bianchi2023easily,friedrich2023fair}. For example, T2I models tend to exclusively generate white, male-appearing people for prompts like `\textit{a photo of a CEO}.' 
Previous research has found no clear correlation between the bias in the distribution of training images and model outputs \cite{friedrich2023fair}. However, some works suggest a connection between a model's bias and the occurrence of gender-connoted terms in the training captions \cite{seshadri2024thebias}. 

\textbf{\texttt{T3\!.\!1} Gender occupation bias correlations.}
We consider the gender distribution in the training images and different caption datasets for various occupations. Subsequently, we compare these ratios with the generation bias of the trained models to identify potential correlations. 

\section{Training and Evaluation: Protocols \& Data}

Before exploring the details of our experimental setup, let us define the goals and scope of this work. 
We aim to provide a structured analysis of important design choices when training T2I models that have been underexplored. Consequently, the goal of the experiments conducted was not to train a new state-of-the-art model. Instead, this analysis furthers the scientific understanding of a crucial aspect of T2I training. 
To ensure that we solely measure the influence of different synthetic texts, we train all checkpoints and baselines using the same architecture, training setup, and training images in a fixed order. Consequently, this setup allows us to only ablate over the associated captions.

\begin{figure*}[t]
    \begin{subfigure}[b]{0.65\textwidth}
         \centering
     \includegraphics[width=.9\linewidth]{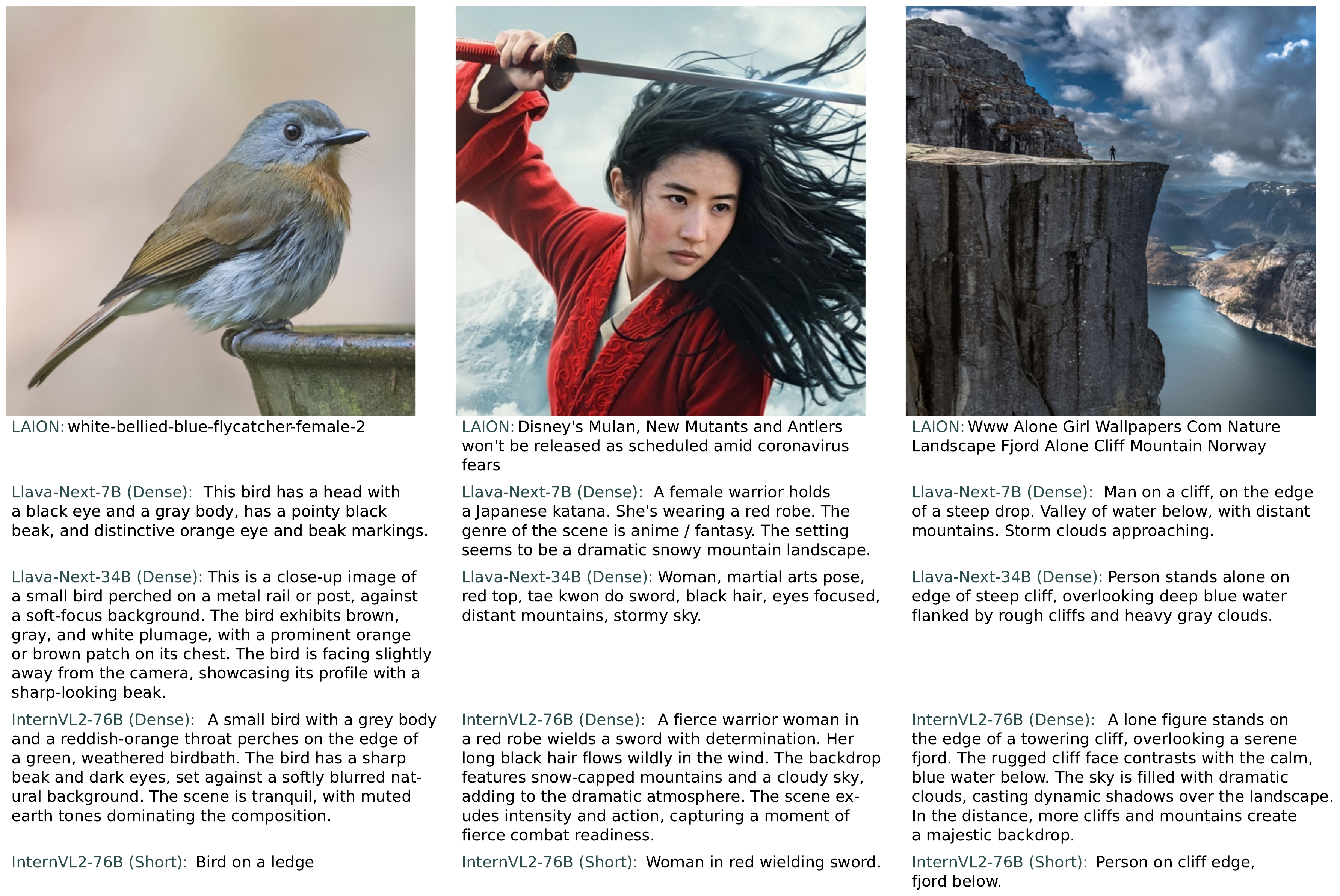}
    \caption{Image and caption examples from our training set.}
     \end{subfigure}
     \hfill
     \begin{subfigure}[b]{0.3\textwidth}
         \centering
         \includegraphics[width=\textwidth]{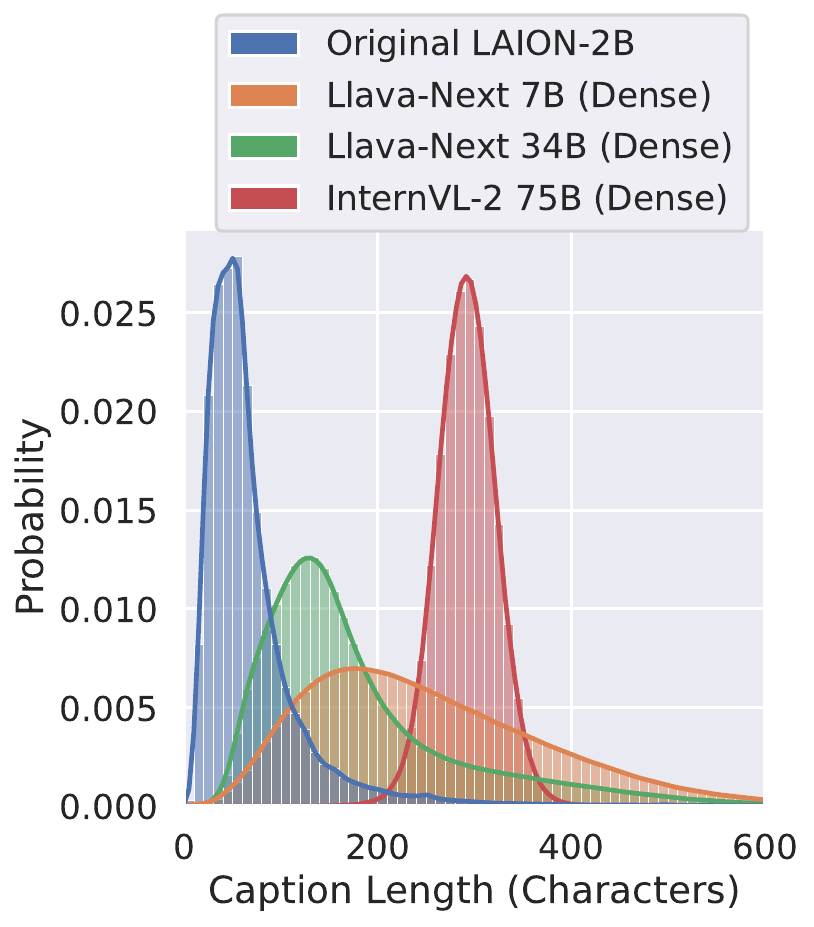}
          \caption{Caption length distribution over 1M image training set.}
          \label{fig:caption_density}
     \end{subfigure}
     \vskip -.25em
    \caption{Examples of different caption distributions for training images. Llava models struggle to produce highly descriptive captions around a target length. They are also more likely to include hallucinations. Conversely, InternVL-2 76B can easily be instructed to generate descriptive captions of different lengths. }
    \label{fig:caption_statistics}
          \vskip -.75em
\end{figure*}

\subsection{Model and Training Setup}
We decided to base our experiments on the original Stable Diffusion (SD) architecture \cite{rombach2022High}. The SD architecture has a proven track record in academic research and is thus well understood by the community. Additionally, the comparatively small model size provides a good trade-off between the required compute while still producing images of good quality. 
We ensure that all captions and prompts used in our experiment fit SD's maximum token length of 77. 

In general, large-scale ablation studies like ours come with significant computational requirements. Arguably, these limitations explain why prior literature has not yet conducted a structured investigation of synthetic captions in T2I training despite its importance to the field. Recognizing these challenges, we opted for an experimental setup that reduces compute costs to a manageable level while providing meaningful and generalizable insights. Intuitively, the early stages of training a T2I model are responsible for establishing basic capabilities in generating coherent images and T2I alignment. More nuanced improvements and differences will likely emerge at later training stages \cite{balaji2022ediffi,li2022mvptr}.

Following this argument, we fix this initial training phase for every experiment, allowing us to reuse a once-trained checkpoint for all ablations. For increased scientific transparency, we use the publicly available SDv1.1 to initialize all subsequent training runs. 
In our experiments, we rewarm the learning rate to 1e-4 over 1000 steps while maintaining the global batch size of 2048 and image resolution of 512x512. In line with the training setup of later SD checkpoints, we drop the text-conditioning for 10\% of samples to improve classifier-free guidance \cite{ho2022classifier}. 
All training runs were conducted on A100-SXM4-80GB using bf16 precision and the AdamW optimizer \cite{loshchilov2018decoupled}. 

Unless otherwise stated, all models are trained for four epochs on 1 Million images. We empirically determined four epochs to be the meaningful cut-off points at which most experiments exhibit an initial plateau in performance improvements. In Fig.~\ref{fig:continual_examples}, we provide qualitative examples highlighting the validity of our continual pre-training setup. For one, the final models yield significant improvements over the baseline SDv1.1 checkpoint in key metrics such as aesthetics and T2I alignment. Further, we observe meaningful differences in the downstream performance of different captioning setups. 

We argue that insights from our experiments can be extrapolated to different model sizes, architectures, and compute budgets. 
This assumption is well grounded in prior studies on scaling laws and architecture impact. Small experiments on the impact of specific data compositions reliably predict behavior in larger models \cite{li2024scalability, magnusson2025datadecide}. Similarly, different machine learning architectures converge to similar performance levels when trained on identical datasets \cite{kim2024transformers, gut2022benchmarking}. Consequently, our paper's observations and recommendations generally apply to text-to-image models. The proposed setup provides a good balance between extensive, meaningful ablations and keeping computational costs at a reasonable level.

\subsection{Training Data}

\textbf{Images} We sourced the training images from LAION Aesthetics and subsequently outlined the data processing steps leading up to our 1 Million sample subset. First, we limit ourselves to highly aesthetic images with decent resolution. Consequently, we only included images with an aesthetics score greater than six and a minimum resolution of 512p. Before downloading, we filtered all images for watermarks and removed potential NSFW and CSAM material before downloading using PhotoDNA\footnote{\tiny\url{https://www.microsoft.com/en-us/photodna/?oneroute=true}}, Thorne\footnote{\tiny\url{https://www.thorn.org/solutions/victim-identification/}}, and TheHive Nudity\footnote{\tiny\url{https://docs.thehive.ai/docs/visual-content-moderation}} classifier. Additionally, we removed exact duplicates of images based on a perceptual hashing algorithm. Lastly, we cropped all images to 512x512 resolution using a `smart' cropping algorithm that aims to center foreground objects. Again, all experiments use the same training images and only differ in the respective training captions. 

\begin{figure*}[t]
   \centering
   \includegraphics[width=.9\textwidth]{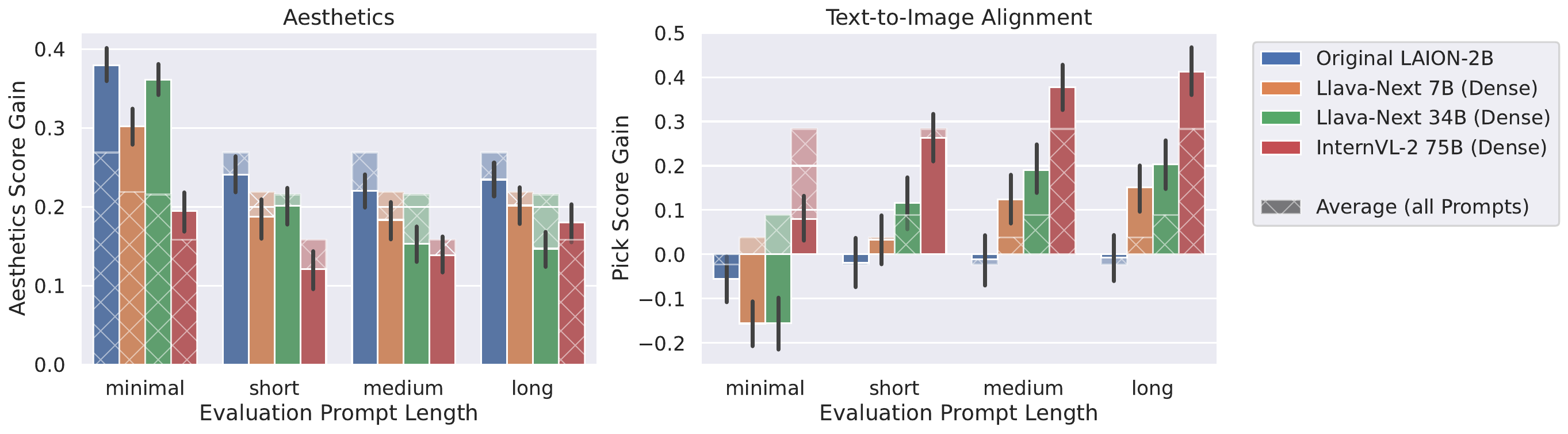}
   \vskip -.25em
    \caption{Comparison of T2I performance when trained on different (synthetic) captions. Scores are shown as the improvement over the SDv1.1 base model. High-quality captions yield better prompt-following capabilities but generate less aesthetic images. }
    \label{fig:model_comp}
          \vskip -.75em
\end{figure*}

\textbf{Captions} We retained the original text captions from LAION 2B as a baseline for subsequent experiments. 
For each experiment, T1.1 - T3.1, we sample dedicated captions corresponding to the tested hypothesis. We make all captions generated in our experiments publicly available.\footnote{\tiny\url{https://huggingface.co/datasets/AIML-TUDA/t2i-diversity-captions}} 
To reduce compute requirements when generating textual descriptions, 
Practitioners will often input downsampled images to the VLM. In App.~\ref{app:resolution}, we compare the downstream performance of models whose training captions only differ in the image resolution used at caption generation. Generating the captions on full-resolution images yields slightly better T2I models. 
However, the differences are small and within the calculated confidence intervals. Consequently, using lower-resolution images during captioning is a valid strategy to reduce computational requirements with only minimal tradeoffs. For all subsequent experiments, we, therefore, use a slightly reduced resolution of 256p when generating captions. 

\subsection{Evaluation Data}


The final component in our evaluation pipeline is the set of prompts used to generate evaluation images. One of the key aspects in the later analysis is performance differences over various levels of prompt length and complexity. To eliminate confounding factors in the evaluation set, each prompt is included at four different levels of descriptiveness but with one shared subject.  We started with over 1000 \texttt{minimal} prompts from drawbench \cite{saharia2022photorealistic} and parti-prompts \cite{yu2022scaling} that are less than ten tokens long. Using GPT-4o, we then add three increasingly complex versions of each prompt: \texttt{short}, \texttt{medium}, and \texttt{long}. 
We depict examples of our prompt set and generated images in Fig.~\ref{fig:eval_prompts}. We make this set of prompts publicly available for other researchers to use\footnote{\tiny\url{https://huggingface.co/datasets/AIML-TUDA/t2i-diversity-evalprompts}}. During evaluation, we compute ten images per prompt, which results in a comprehensive evaluation set of over 40,000 images per checkpoint. In the main body of the paper, we report benchmark improvements over the SDv1.1 base model instead of absolute values to eliminate potentially confounding factors further.
Additionally, we corroborate the main finding of this paper on GenAI-Bench \cite{li2024genaibench} using the proposed VQA-Score (cf. App.~\ref{app:genai_bench}).

For gender bias evaluation, we chose 150 occupation prompts \cite{friedrich2023fair} and generated 100 images each.

\subsection{Evaluation Metrics}

Performing meaningful evaluations of T2I models is notoriously challenging. For one, overall model performance cannot be reduced to one metric but includes multiple aspects like image aesthetics, T2I alignment, compositionality capabilities, and sample diversity. 
Further, traditional evaluation metrics such as FID \cite{NIPS2017_8a1d6947} or CLIP scores \cite{hessel2021clipscore} have been shown to correlate poorly with human preference \cite{Kirstain2023PickaPicAO,parmar2022aliased}.

Instead, we use stronger metrics that are more aligned with human preferences.
To evaluate output aesthetics, we use the LAION Aesthetics Classifier v2\footnote{\tiny\url{https://laion.ai/blog/laion-aesthetics/\#laion-aesthetics-v2}}. Additionally, we rely on PickScore \cite{Kirstain2023PickaPicAO}, which has been demonstrated to approximate human preferences for T2I alignment and overall scene composition. Lastly, we judge a model's inter-sample diversity through pair-wise LPIPS distance \cite{zhang2018perceptual} over multiple images generated from the same prompt. For gender bias, we classify images, similar to \cite{friedrich2023fair,friedrich2024multilingual}, with FairFace \cite{karkkainenfairface} and compare gender ratios.


    

\begin{figure*}[t!]
\begin{subfigure}[t]{0.75\textwidth}
         \centering
    \includegraphics[width=.9\linewidth]{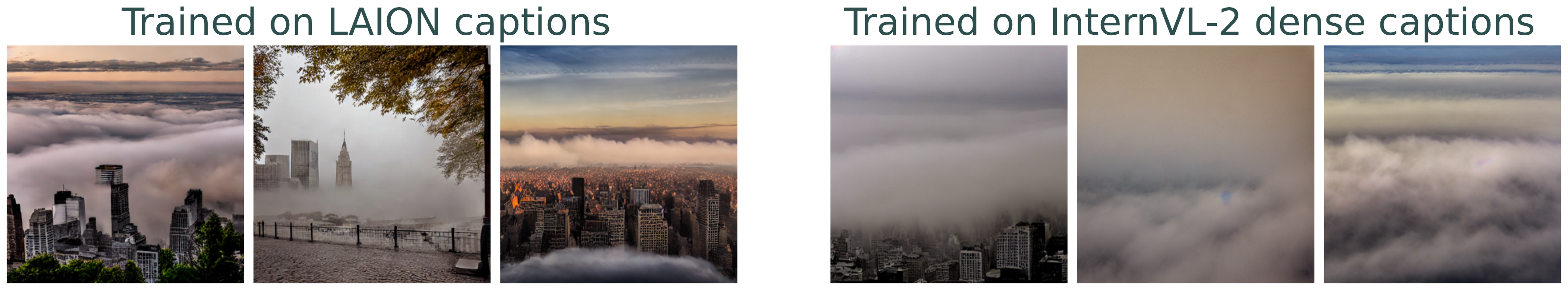}
    \caption{Images generated for prompt `\textit{fog rolling into New York City}'. The model trained on long, dense captions yields less diverse images.}
     \end{subfigure}
     \hfill
     \begin{subfigure}[t]{0.22\textwidth}
         \centering
         \includegraphics[width=\textwidth]{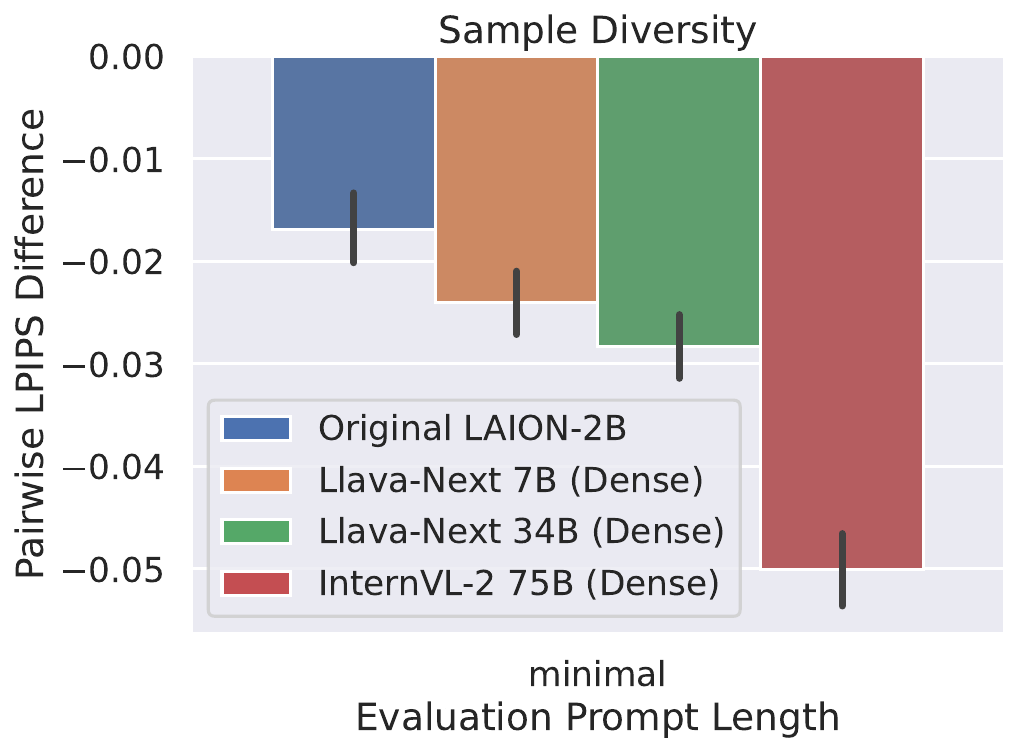}
          \caption{Pairwise LPIPS score between images for the same prompt.}
     \end{subfigure}   
    \vskip -.3em
    \caption{Training on long, dense captions leads to \textit{bland} images generated for minimal text prompts.}
    \label{fig:blandness}
          \vskip -.75em
\end{figure*}


\section{Results}\label{sec:exp1}
With our methodology established, we now empirically test each hypothesis \texttt{H} from Sec.~\ref{sec:hyp}. Based on the results from each experiment \texttt{T}, we derive insights and recommendations for T2I training.

\subsection{\textbf{\texttt{H1}}: Trade-offs of Long, Dense Captions}

For \texttt{T1\!.\!1}, we sample a dense, long caption for each training image from three pre-trained VLMs of increasing sizes and capabilities: LLava-Next-7B, Llava-Next-34B \cite{li2024llava}, and InternVL2-Llama3-76B \cite{chen2024far}. In Fig.~\ref{fig:caption_statistics}, we compare the density of captions generated by all three models with the baseline captions from LAION. The original captions from the LAION dataset are very minimalistic on average. 
Further, both Llava models struggle to generate long captions consistently, whereas InternVL2 yields descriptions of consistent length.
Additionally, the qualitative examples in Fig.~\ref{fig:caption_statistics} highlight that these web-crawled captions are considerably noisy and oftentimes not related to the image content at all. Further, we found the LLava models to be more likely to introduce hallucinations than the larger InternVL2 model (e.g., colors of the bird). We describe the exact sampling setup in further detail in App.~\ref{app:exp_details}.

\textbf{Synthetic captions improve T2I alingment.}
We depict the evaluation results of checkpoints trained on these captions in Fig.~\ref{fig:model_comp}. Since there are a lot of insights to unpack from this Figure, let us dissect it step by step. 
First, we observe that training captions from stronger models produce better T2I alignment. The results indicate clear improvements from the noisy LAION captions to those generated by LLava-Next 7B and 34B, with InternVL-2 76B strongly outperforming all other setups. 
This performance difference can be attributed to less hallucinatory and more accurate descriptions from stronger VLMs. Thus, this enables the T2I model to learn the right associations between all aspects of an image and its natural language description much better.

\begin{figure*}[t]
\centering
   \includegraphics[height=9em]{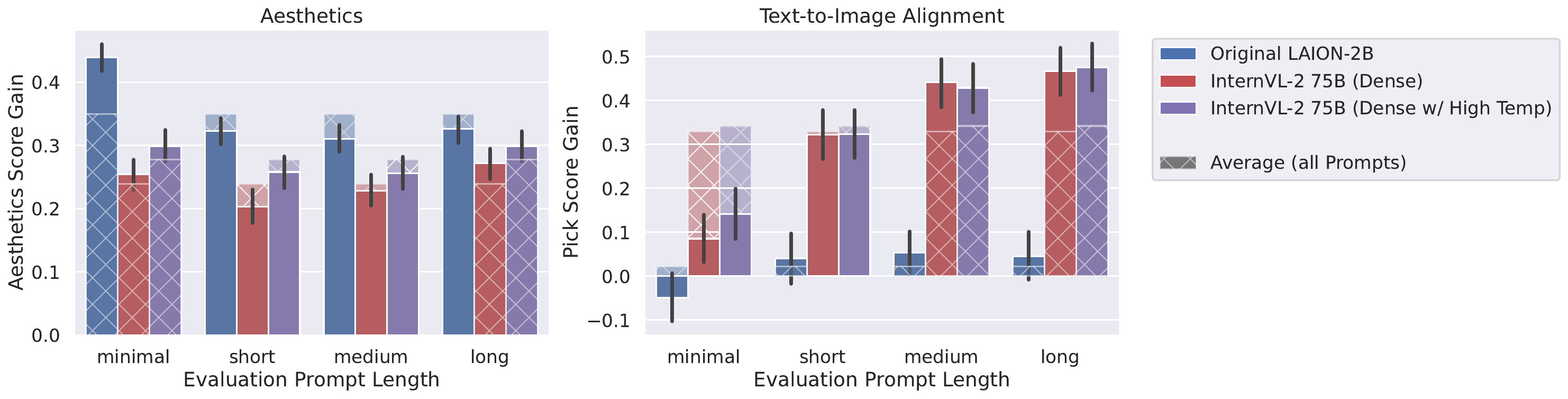}%
   \includegraphics[height=9em]{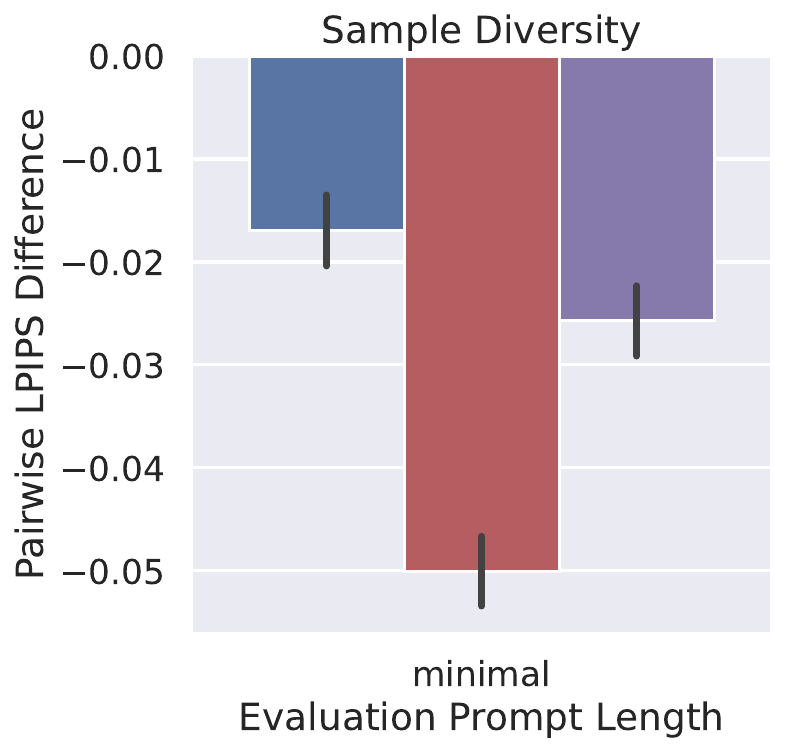}
   \vskip -.3em
    \caption{Increasing caption diversity through higher sampling temperature improves T2I performance. Scores are shown as the improvement over the SDv1.1 base model.}
    \label{fig:temp}
          \vskip -.75em
\end{figure*}

\begin{figure*}[b]
\centering
   \includegraphics[height=9em]{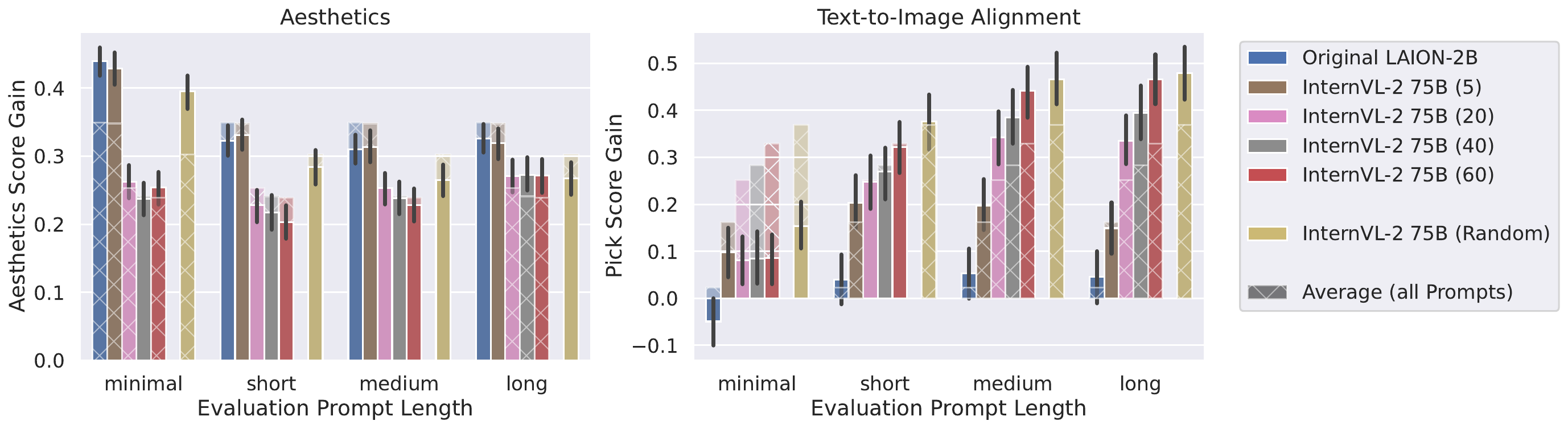}%
\includegraphics[height=9em]{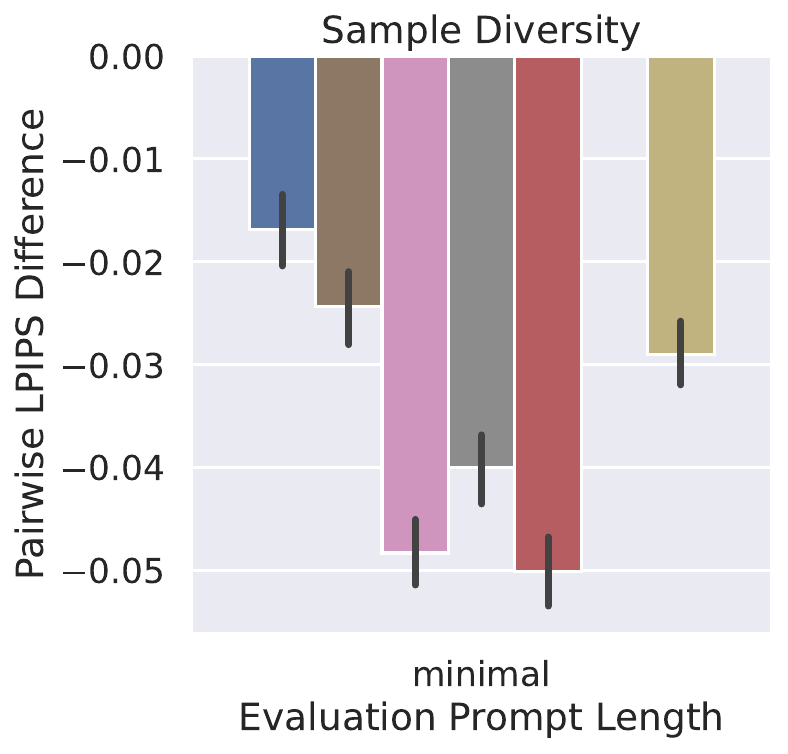}
   \vskip -.25em
    \caption{Comparison of T2I model performance when trained on different length captions. Fixed-length distributions lead to a trade-off between output aesthetics and text following. Randomizing the caption length at training results in the overall best model.  Scores are shown as the improvement over the SDv1.1 base model.}
    \label{fig:length}
          \vskip -1em
\end{figure*}

Next, the evaluation indicates that training on dense captions results in better alignment when using long prompts at inference. In fact, the checkpoints using LLava captions even decrease T2I alignment for minimal prompts through training and achieve worse performance than the noisy LAION captions.
Intuitively, these prompts are pushed out of distribution when only training on long captions. 
Moreover, training with high-quality InternVL2 captions substantially enhances text alignment across all prompt lengths, reinforcing the benefits of using synthetic captions.

\textbf{Output aesthetics \& instruction following as competing objectives.}
Turning to image aesthetics, we observe that all setups markedly enhance the visual appeal of the outputs—a result that aligns with expectations, as the images used in our continual pre-training are inherently high in aesthetic quality. However, notable differences emerge in the aesthetic quality of outputs across different training setups. This finding is somewhat unexpected, given that all models were trained on the identical set of images, and the captions alone would not be anticipated to impact aesthetics directly. Interestingly, the setup with the highest caption noise (i.e., the original LAION captions) produces the most aesthetically pleasing results overall. Further, the observed trade-off remains consistent throughout other experiments in this paper, demonstrating a general trend.

The results indicate that text alignment and aesthetic capabilities of a T2I model seem to be competing objectives. Consequently, models with limited capacity will often optimize for only one of the two tasks. This competition results from the rather implicit training objective of reconstructing the original image. Downstream capabilities like image aesthetics or prompt alignment are not modeled explicitly during training.  When presented with noisy captions, the strongest training signal can be derived from capturing the images' aesthetics. Conversely, highly descriptive and semantically rich captions push the model toward learning meaningful representations of the presented concepts. These orthogonal objectives could be (partially) addressed by increasing the model's capacity by scaling up its parameter count. Nonetheless, our subsequent experiments in Sec.~\ref{sec:diversification} explore different captioning techniques to optimize the observed tradeoff for small models. 

\textbf{Out-off distribution issue for minimal generation prompts.}
Beyond this general competition in objectives, Fig.~\ref{fig:model_comp} also shows that the difference in output aesthetics between the LAION and InternVL model is considerably higher for \texttt{minimal} prompts. Upon manual investigations, we identified a certain `blandness' in the InternVL model's images as the key factor in that performance gap. The generated images often have generic gray backgrounds and little diversity between samples.
We provide some qualitative examples in Fig.~\ref{fig:blandness} along with an empirical evaluation supporting these observations. This effect comes back to the InternVL model being trained on captions describing every image aspect. When presented with \texttt{minimal} prompts at inference, the model tends to sample the most generic choices for aspects not mentioned specifically, such as grey backgrounds. We can measure this effect by comparing the inter-sample diversity of multiple images generated with the same prompt. As shown in Fig.~\ref{fig:blandness}, the pairwise LPIPS distance between sampled images is indeed drastically lower for images sampled from the InternVL model.

\begin{tcolorbox}[leftrule=1.5mm,top=0.8mm,bottom=0.5mm,title=Key Insights]
\begin{itemize}
    \item Training on long, dense captions is not unequivocally better then short captions
    \item Dense caption training improves prompt following at the cost of other aspects (e.g. aesthetics)
    \item Performance loss is particularly strong for short prompts
\end{itemize}
\end{tcolorbox}

\begin{figure*}[t]
\centering
    \includegraphics[height=9em]{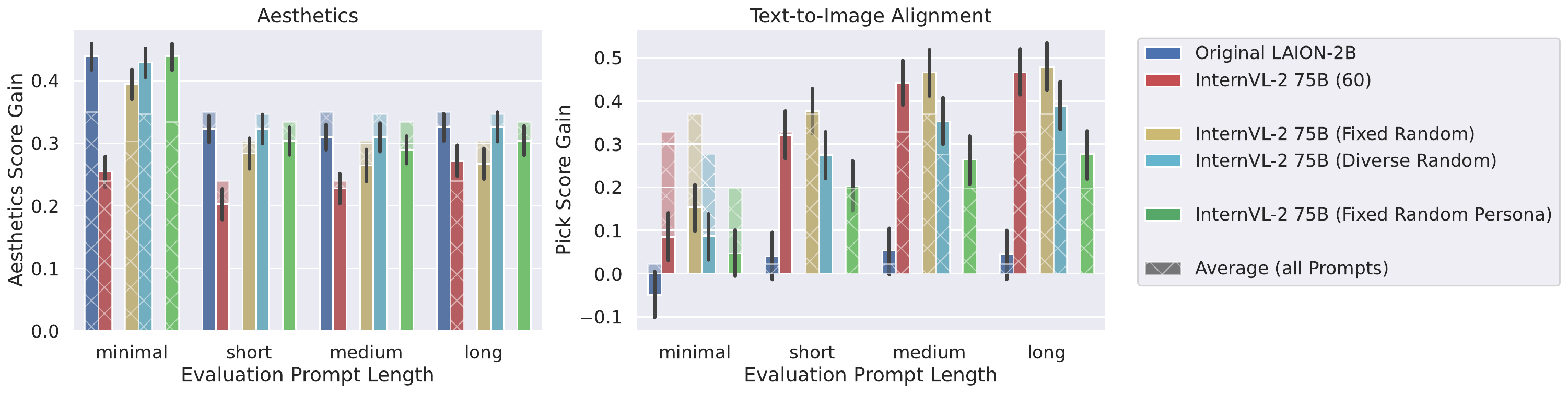}%
    \includegraphics[height=9em]{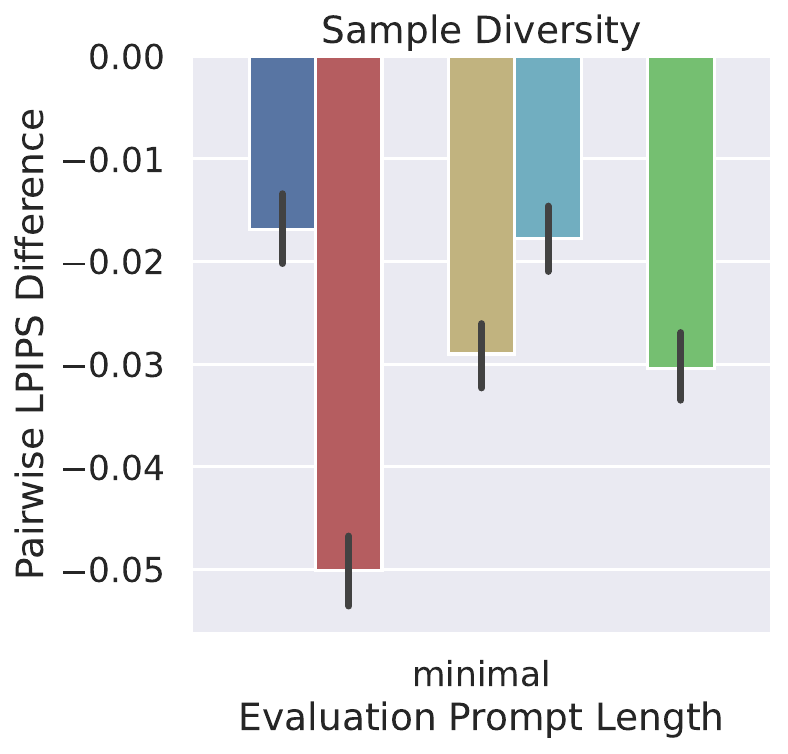}   
    \vskip -.3em
    \caption{Comparison of performance impact using more sophisticated captioning setups. Varying captions for each epoch or employing personas during caption generation does not improve T2I performance. Scores are shown as the improvement over the SDv1.1 base model.}
    \label{fig:diverse_and_persona}
      \vskip -.75em
\end{figure*}

\subsection{\textbf{\texttt{H2}}: The Right Diversity Will Help You}\label{sec:diversification}
Based on the insights from the previous Section, we now increase diversity in the synthetic captioning pipeline. We aim to maintain the strong capabilities in scene composition and text alignment produced by the high-quality, dense captions of InternVL2. At the same time, we want to address the limitations we observed with respect to overall image aesthetics and diversity.

\textbf{\texttt{T2\!.\!1} Increasing sampling temperatures reduces the aesthetics gap.}
Through manual inspection, we chose a sampling temperature of $1.5$, which provided significantly more diverse captions while maintaining accurate descriptions. 
We show the downstream performance of the model trained on these captions in Fig.~\ref{fig:temp}. The results show a small but meaningful improvement in image quality and sample diversity through this diversification. Importantly, that improvement does not come at the cost of the model's text-alignment capabilities.

\textbf{\texttt{T2\!.\!2} Diversifying caption length improves aesthetics and alignment.} 
In Sec.~\ref{sec:exp1}, we demonstrated that only training on long, descriptive captions pushes short prompts out of distribution. 
Consequently, we now explore the influence of caption length on the trained T2I model.
We found InternVL2 surprisingly reliable in that regard when simply stating the target length of an image description in the prompt. Additionally, we trained a checkpoint on a caption of randomized length for each training image. The performance of all versions is depicted in Fig.~\ref{fig:length}. 

Once again, let us dissect the results step by step. First, we can see that the model trained on minimal, synthetic captions results in comparable image aesthetics as the one using LAION captions. At the same time, these more accurate scene descriptions result in better prompt following on the observed level of complexity. For minimal prompts, this model remains competitive with the one trained on dense captions but underperforms for increasingly complex prompts. Conversely, the longest captions result in the best or equal prompt following capabilities across all prompt complexities at inference. 

Importantly, the model trained on random-length captions no longer trade-off text alignment with output aesthetics. Not only does that checkpoint achieve the best pick scores across the board, but it also significantly closes the gap in aesthetics score, especially on the crucial  \texttt{minimal} prompts. At the same time, randomizing training caption length also reduces the drop in inter-sample diversity observed for longer captions.




\textbf{\texttt{T2\!.\!3}: Inter Epoch Diversity reinforces aesthetics-alignment tradeoff.}
We test the hypothesis of inter-epoch diversity in two scenarios. First, we extend our prior setup from \texttt{T2\!.\!2} by presenting a different caption of random length at each epoch for each image. We compare the downstream performance of the model trained on those captions in Fig.~\ref{fig:diverse_and_persona}. Overall, we observe that while this approach slightly improves output aesthetics, it compromises the model’s ability to follow text instructions effectively, providing no clear advantage. Instead, it reflects a different point on the previously demonstrated trade-off spectrum.

We also conducted a rigorous evaluation of data-constrained training scenarios. Interestingly, our findings indicate that DM training remains robust despite limited data, with no observable benefit from inter-epoch caption variation. Details of this analysis are provided in App.~\ref{app:inter_epoch}.

\textbf{\texttt{T2\!.\!4}: Persona diversity drastically decreases prompt-following capabilities.}
We constructed 20 different personas, each with instructions on image aspects the respective captions should capture. 
Using a two-stage setup, we first tasked the VLM to asses which personas were best suited to describe a given image, then sampled captions of varying lengths tailored to each persona. We provide persona prompts and examples in App.~\ref{app:personas}.

When manually inspecting, the captions generated by this process, 
we observed strong stylistic variations
However, the empirical evaluation in Fig.~\ref{fig:diverse_and_persona} does not support this first intuition. Instead, we observe a drastic decrease in overall prompt-following capabilities. While the idea of using different personas during caption sampling remains interesting on paper, measurable benefits remain to be established in future work.  



\begin{figure}[t]
    \centering
    \includegraphics[width=.8\linewidth]{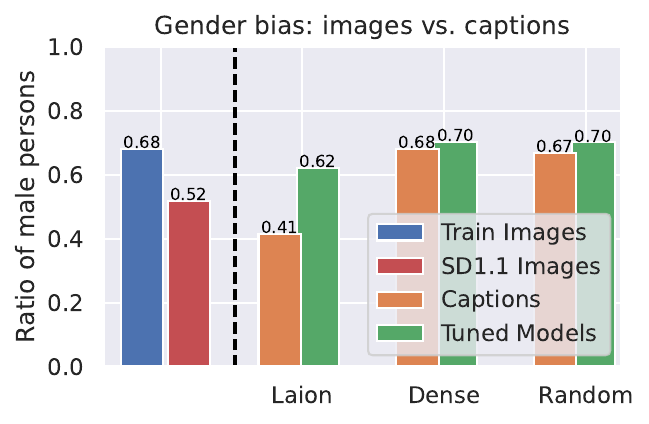}
    \vskip -.5em
    \caption{Comparing gender ratios in generated images with dataset distributions.}
    \label{fig:occupation-bias-combined}
      \vskip -.75em
\end{figure}


\begin{tcolorbox}[leftrule=1.5mm,top=0.8mm,bottom=0.5mm,title=Key Insights]
\begin{itemize}
    \item Diversifying captions eliminated aesthetics vs. prompt-following tradoff
    \item \textbf{Recommendation:} Randomly sample length/density of training captions
    \item We observed no improvements when diversifying training captions over epochs or through persona sampling
\end{itemize}
\end{tcolorbox}

\subsection{\textbf{\texttt{H3}}: Captions Strongly Influence Model Bias.}
Before concluding, we investigate how bias in training captions affects the trained T2I model. We focus on gender bias, as it is a well-studied area for T2I models and is deeply embedded in language and grammar (e.g., ``a woman and her son with his toy'').

\textbf{\texttt{T3\!.\!1}: Caption distributions influences gender bias.} 
Fig.~\ref{fig:occupation-bias-combined} shows the average gender ratio across occupations. 
We observe a remarkable shift in bias between the baseline SDv1.1 and tuned models (red vs.~green bars). 
Importantly, this bias shift varies considerably despite all checkpoints being trained on the same images. 
When using dense or random-length captions sampled from InternVL2, the gender ratios closely follow the image distribution (blue bar), leading to aligned output biases across models. In contrast, training with original Laion captions results in a less pronounced shift, as these captions refer to males less frequently. This highlights the strong influence of training captions on downstream model bias.

Yet, the model trained on Laion captions does not strictly adhere to the caption distribution when generating images, suggesting a more complex relationship between captions and model bias. This contrasts previous work \cite{seshadri2024thebias} and suggests that multiple factors must be considered and more research is needed. Nonetheless, using well-crafted captioning models helps align output biases with the image distribution.
We provide further bias analysis details in App.~\ref{app:bias}. 

\begin{tcolorbox}[leftrule=1.5mm,top=0.8mm,bottom=0.5mm,title=Key Insights]
\begin{itemize}
    \item Training caption distribution influences model biases 
    \item Exact interaction between distributions are complex
\end{itemize}
\end{tcolorbox}

\section{Limitations}
Before concluding, let us touch on potential limitations of our work. 
The employed continual pre-training setup, instead of training all ablations from scratch, could theoretically introduce a confounding factor. However, the key finding of this paper can be drawn from both absolute benchmark performance and relative gains. Additionally, we still observed significant improvements in the types of data the base model was trained on, i.e., short captions. Consequently, we believe any confounding to be limited. 

We did not explicitly ensure that the generated captions are free from hallucinations. However, InternVL2 performs well on Hallusionbench \cite{guan2024hallusionbench}, a benchmark designed to uncover hallucination issues in VLMs. Additionally, we manually reviewed a subset of generated captions for hallucinations and found no incorrect scene descriptions for sampling temperatures below $1.7$ (which is true for all our setups).

Lastly, we did not conduct ablations over different T2I architectures of our model sizes or fine-tune captioning models, which we leave for future work. 

\section{Conclusion}

In this study, we explored key design choices in synthetic captioning for training T2I models. Our findings show that while high-quality, descriptive captions improve T2I alignment, they often compromise the aesthetic quality and variety of outputs, especially for minimal prompts. Conversely, noisy captions, despite their lack of semantic precision, yield more visually appealing and diverse outputs, highlighting an intrinsic tension between model aesthetics and semantic fidelity.
We also demonstrated that diversifying caption characteristics, such as length and density, can mitigate these trade-offs. Random-length captions, in particular, are effective in balancing alignment with aesthetic appeal and inter-sample diversity. 
Our work provides avenues to refine T2I training protocols further and offers practical guidelines for synthetic captioning. Future work should investigate further nuances in caption diversity, including contextual or persona-based captioning.
This research contributes to developing more robust and adaptable T2I models by advancing our understanding of \mbox{captioning strategies.}

\section*{Acknowledgments}
e acknowledge support of the hessian.AI Innovation Lab
(funded by the Hessian Ministry for Digital Strategy and
Innovation), the hessian.AISC Service Center (funded by
the Federal Ministry of Education and Research, BMBF,
grant No 01IS22091), and the Centre for European Re-
search in Trusted AI (CERTAIN). Further, this work benefited from the ICT-48 Network of AI Research Excellence
Center “TAILOR” (EU Horizon 2020, GA No 952215),
the Hessian research priority program LOEWE within the
project WhiteBox, the HMWK cluster projects “Adaptive
Mind” and “Third Wave of AI”, and from the NHR4CES.
{
    \small
    \bibliographystyle{ieeenat_fullname_iccv}
    \bibliography{main}
}

\clearpage
\setcounter{page}{1}
\maketitlesupplementary
\setcounter{section}{0}
\renewcommand{\thesection}{\Alph{section}}

\section{Further Experimental Details}\label{app:exp_details}
We here provide further experimental details not eluded to in the main body of the paper. 

\subsection{Synthetic Caption Sampling}\label{app:exp_details_sampling}
For the Llava-Next models, we used the sglang \cite{zheng2024sglang} as the inference framework, whereas we relied on lmdeploy \cite{2023lmdeploy}for InternVL2-LLama3-76B. Unless expressly stated otherwise, we used the default sampling hyperparameters at a batch size of 128 and image input resolution of 256p.
Through careful prompt engineering, we decided to use the following captioning prompt. 

\texttt{Give a description of the objects and scene in the image as if the description can be used to prompt a text-to-image generator model to generate images. Do not start the description with "The image". Do not exceed $<$X$>$ words.}

To generate long, dense captions, for example, we would set \texttt{$<$X$>$} to 60.

\begin{figure*}[t]
\centering
    \includegraphics[width=.9\linewidth]{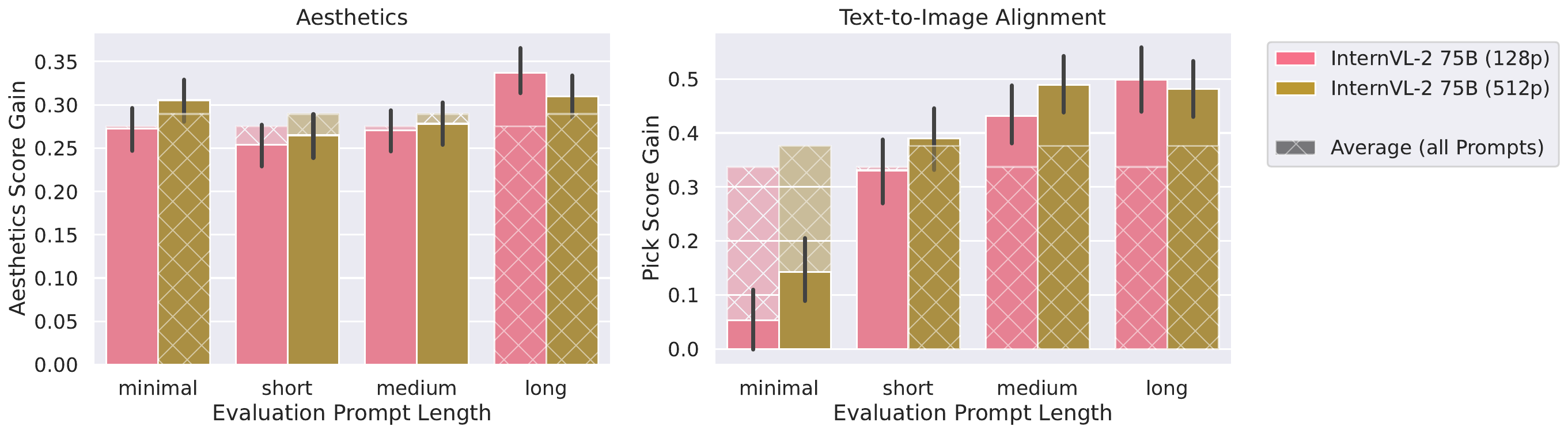}

    \caption{Comparison of training on captions sampled for low or high-resolution images. Decreasing image resolution for the VLM captioning can be a worthwhile trade-off. Scores are shown as the improvement over the SDv1.1 base model.}
    \label{fig:resolution}
\end{figure*}
\begin{figure*}[t]
\centering
    \includegraphics[width=.9\linewidth]{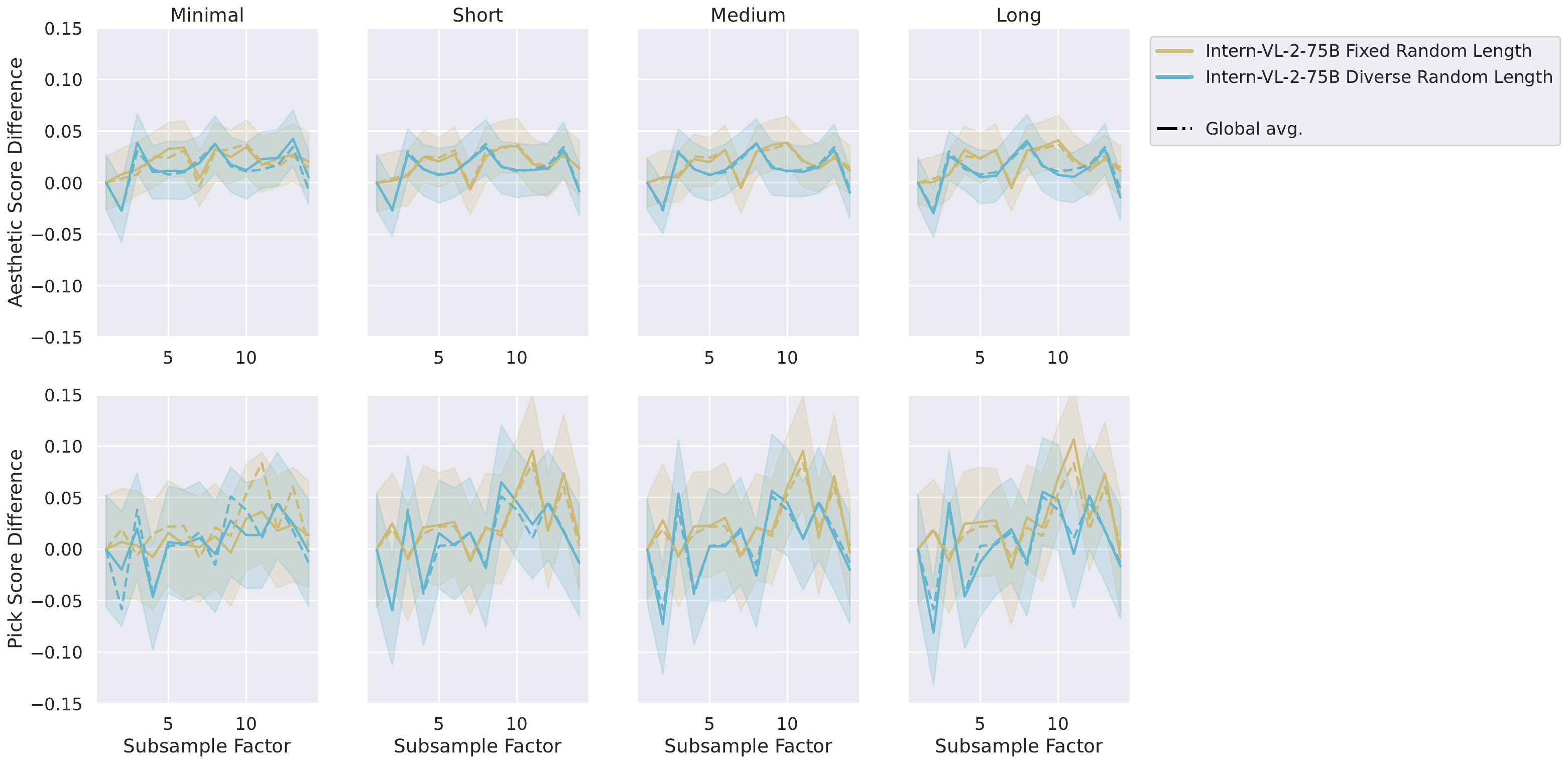}

    \caption{Comparison of training on diversifying captions over epochs in data-constrained environments. Subsampling factor refers to the subset size of training images and respective epochs. Scores are depicted as difference to the model trained on all 1M images.}
    \label{fig:subsampling}
\end{figure*}

\begin{figure*}[t]
\centering
    \includegraphics[width=.9\linewidth]{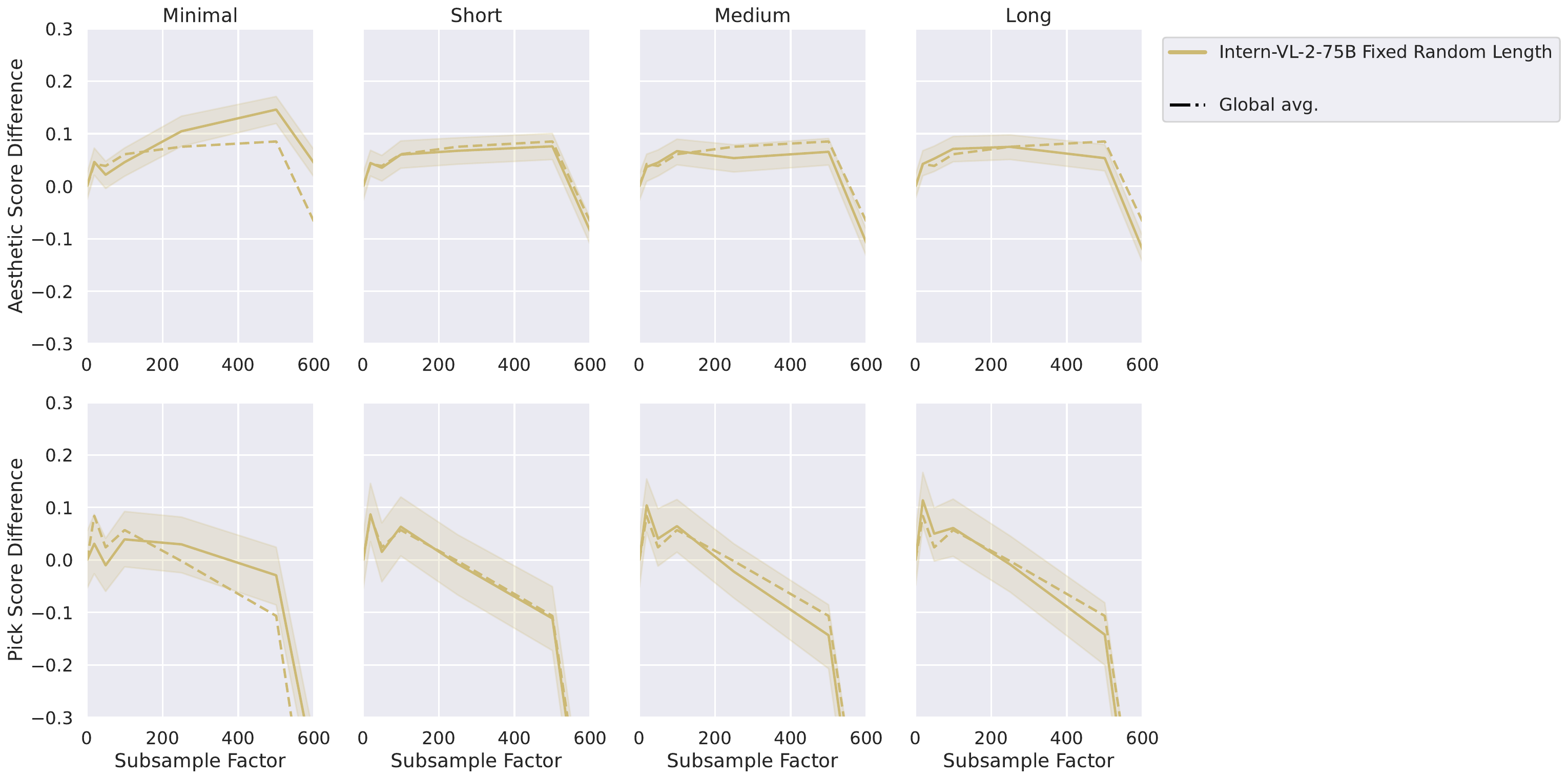}

    \caption{Diffusion models behave suprisingly robust in data-constrained training. Subsampling factor refers to the subset size of training images and respective epochs. Scores are depicted as difference to the model trained on all 1M images.}
    \label{fig:subsampling2}
\end{figure*}

\subsection{Training Framework}
We used an adapted version of the \texttt{accelerate} training script for Stable Diffusion provided by HuggingFace.
The script is readily available online as part of the \texttt{diffusers} library\footnote{\tiny\url{https://github.com/huggingface/diffusers/blob/main/examples/text_to_image/train_text_to_image.py}}. 

\section{Image Resolution}\label{app:resolution}
To reduce compute requirements when generating textual descriptions, 
Practitioners will often input downsampled images to the VLM. In Fig.~\ref{fig:resolution}, we compare the downstream performance of models whose training captions only differ in the image resolution used at caption generation. Generating the captions on full-resolution images yields slightly better T2I models. 
However, the differences are small and within the calculated confidence intervals. Consequently, using lower-resolution images during captioning is a valid strategy to reduce computational requirements with only minimal tradeoffs. For all subsequent experiments, we, therefore, use a slightly reduced resolution of 256p when generating captions. 

\definecolor{blue}{HTML}{1F77B4}
\definecolor{red}{HTML}{D62728}
\definecolor{gold}{HTML}{BCBD22}
\definecolor{turquoise}{HTML}{17BECF}
\definecolor{green}{HTML}{6ACC64}
\section{GenAI-Bench Results}\label{app:genai_bench}
\begin{table}[t]
    \centering
    \begin{tabular}{l l c }
     &\textbf{Caption Setup}    &  \textbf{VQA-Score}\\ \hline
     \cellcolor{blue}&Original LAION-2B & 0.6022$^{\pm0.212}$\\
     \cellcolor{red}&InternVL-2 75B (60) & \textbf{0.6581}$^{\pm0.213}$\\
     \cellcolor{gold}&InternVL-2 75B (Fixed Random) & 0.6502$^{\pm0.211}$\\
     \cellcolor{turquoise} &InternVL-2 75B (Diverse Random) &0.6551$^{\pm0.209}$\\
     \cellcolor{green}&InternVL-2 75B (Fixed Random Persona) &0.6311$^{\pm0.212}$\\
    \end{tabular}
    \caption{Evaluation on Gen-AI Bench supports the general findings from the main paper. Superscript numbers indicate the standard deviation. Colors corresponding to the ones in the main body of the paper are included for convenience. }
    \label{tab:genai_bench}
\end{table}

In addition to the evaluation of our own benchmark using pick-score, we corroborate our main findings using GenAI-Bench with the VQA-Score metric \cite{li2024genaibench}. 
Furthermore, we here report the \textit{absolute} scores instead of the difference to the base model. Thus showcasing that our continual learning setup does not confound the analysis conducted. We show the results of our evaluation in Tab.~\ref{tab:genai_bench}.

From these results, we can draw the same conclusions as those in the main body of the paper. 
\begin{enumerate}
    \item Training on long, dense captions results in better prompt following of the downstream text-to-image model
    \item Randomizing training caption length does not adversely affect text alignment while providing the benefits outlined in Sec. \ref{sec:diversification}
    \item We observe no measurable benefit in additionally diversifying captions across epochs
    \item Attempting to make caption diversity explicit through personas results in a performance drop
\end{enumerate}

\section{Inter Epoch Diversity}\label{app:inter_epoch}
In addition to the experiment discussed in Sec.~\ref{sec:diversification}, we conducted a more rigorous evaluation of inter-epoch diversity for data-constrained scenarios. 

\paragraph{Experimental Setup}
For this set of experiments, we assume a fixed training budget of 1M samples. Consequently, training one epoch on our established dataset serves as the baseline for downstream text-to-image performance.

Next, we artificially decrease the number of available images for training by subsampling from the original training set. For example, at a sub-sampling factor of 2x, we sample 500k images on which we would train for two epochs. Consequently, the total number of training steps remains the same.

To ensure that each subsample is representative of the entire training set, we chose not to draw samples randomly. Instead, we build roughly 1.4k image clusters using SigLIP embeddings \cite{zhai2023sigmoid} of all 1M images. We then sample from each cluster based on its entropy. 

We compare two approaches to multi-epoch training for subsampling factors from 2x to 15x. The baseline approach always uses the same random-length caption for each image, whereas the diverse setting shows a different caption at each epoch. 

\paragraph{Results}
We depict the results in Fig.~\ref{fig:subsampling}. First, we observe that text-to-image training is surprisingly stable in this setting.
Even when training on only 66k text-image pairs for 15 epochs, the downstream performance remains similar to training on the full set of 1M images. 
Consequently, we found no benefit in using a different caption at each epoch. 

In fact, we only start to see deterioration in text-to-image alignment when reducing the number of samples below 5000 pairs (cf. Fig.~\ref{fig:subsampling2}). For image aesthetics, the model still produces comparable outputs when trained on only 2000 samples. We argue that the robustness in training can largely be attributed to the fact that we sampled from semantic image clusters. These results further strengthen the recommendation from the main paper regarding training data diversity.

\section{Personas}\label{app:personas}
In the following, we provide further details and examples of the explicit persona sampling discussed in Sec~\ref{sec:diversification}. For any given image, we first prompt the VLM to asses which personalities are suitable for that image. For example, a description from a biologist focusing on flora and fauna would not be particularly helpful if no natural elements were present in the image. 
To that, end we use the following prompt, which we prepend to the 20 persona descriptions outlined in Tabs.~\ref{tab:personas} and \ref{tab:personas_2}.
\texttt{The task is to describe the objects and scene in the image as if the description can be used to prompt a text-to-image generator model to generate images.
For the given roles provide a True or False rating if the role is suitable to give a caption for the image.}

\begin{figure*}[t]
\centering
    \includegraphics[width=\linewidth]{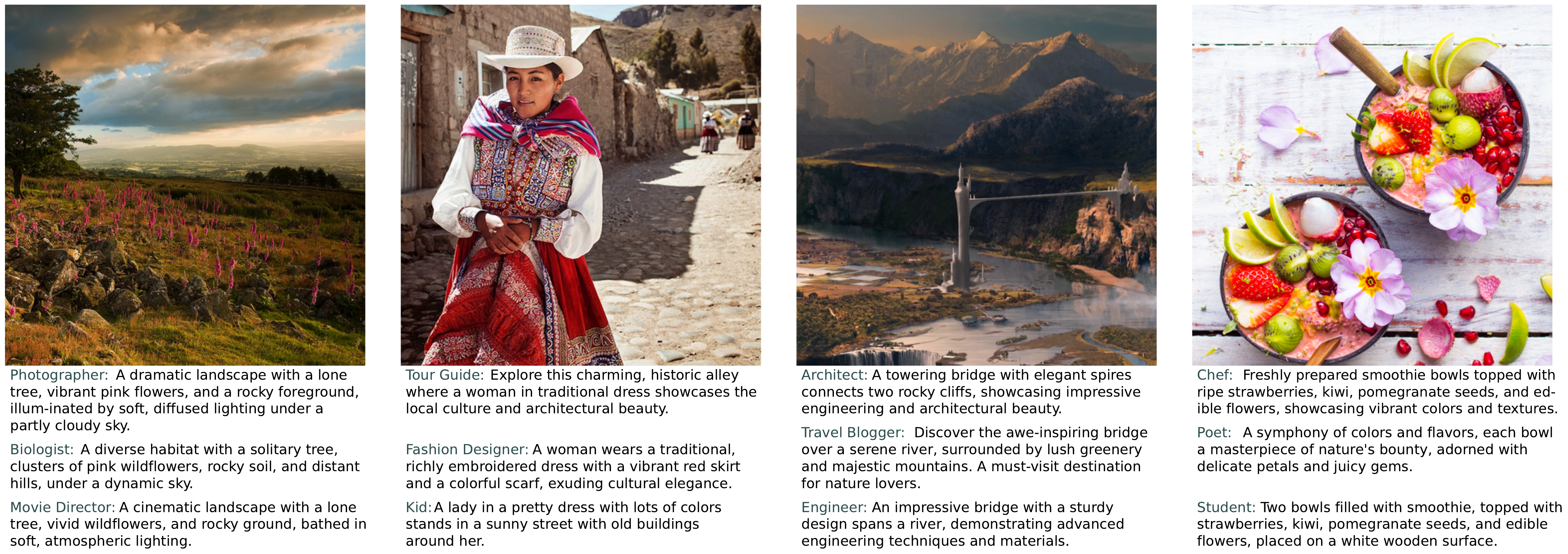}
    \caption{Qualitative examples of diverse persona captions. }
    \label{fig:persona_examples}
\end{figure*}

Subsequently, we prompt the model to generate a response JSON with one description per selected role. Here, we use an instruction prompt similar to the one described in Sec.~\ref{app:exp_details_sampling}. We sampled at least three captions of different lengths for each role, resulting in at least 20 captions per image. We depict some qualitative examples in Fig.~\ref{fig:persona_examples}. The different persona captions provide a lot of diversity on subjective terms like \textit{cinematic}, \textit{vivid}, or \textit{impressive}. In practice, such formulations are often used in prompting text-to-image models and can also be found in GenAI-Bench.

\begin{table*}[t]
 \centering
      \begin{tabular}{p{0.15\linewidth} | p{.85\linewidth}}
     \textbf{Persona}  & \textbf{Description}\\ \hline
Photographer & \texttt{\footnotesize Focuses on technical aspects like lighting, composition, focal points, exposure, and depth of field. The description highlights how these elements work together to create a visually striking or harmonious image.}\\ \hline
Artist & \texttt{\footnotesize Emphasizes emotions, aesthetics, color theory, and the overall atmosphere. This role leans on subjective interpretation, creating a narrative that may evoke feelings or inspire creativity.}\\ \hline
Student & \texttt{\footnotesize Provides a straightforward, observational description of what is depicted, often focusing on what is learned or noticed. The language is usually simple and direct, highlighting key visual elements.}\\ \hline
Kid & \texttt{\footnotesize The description is often simple, imaginative, and filled with curiosity. It tends to focus on vibrant and exciting details, sometimes personifying objects or describing them with enthusiasm.}\\ \hline
Scientist & \texttt{\footnotesize Describes the scene with precision, often breaking down biological, physical, or chemical details. The focus is on facts, measurements, and processes that may not be immediately obvious but are scientifically relevant.}\\ \hline
Historian & \texttt{\footnotesize Provides context around the historical significance of the objects, architecture, or figures in the scene. It often includes dates, origins, cultural relevance, and how the past informs the present.}\\ \hline
Poet & \texttt{\footnotesize Uses metaphorical and symbolic language to evoke emotion and mood. The description is often lyrical and less concerned with technical accuracy, instead aiming to capture an essence or deeper meaning.}\\ \hline
Architect & \texttt{\footnotesize Focuses on the structure, materials, geometry, and spatial design. Descriptions include functional and aesthetic details, often emphasizing how the space interacts with its environment or inhabitants.}\\ \hline
Fashion Designer & \texttt{\footnotesize Describes garments, fabrics, textures, and the fit of clothing on the body. Attention is given to color schemes, trends, and how the overall look conveys style or message.}\\ \hline
Chef & \texttt{\footnotesize Describes food in terms of appearance, texture, and taste, often highlighting the freshness or quality of ingredients. Presentation, plating, and culinary techniques are emphasized.}\\ \hline
Movie Director & \texttt{\footnotesize Provides a cinematic description of the scene, focusing on atmosphere, lighting, framing, and potential narrative. This role often includes emotional undertones or the suggestion of a storyline.}\\ \hline
Tour Guide & \texttt{\footnotesize Offers a guided description, highlighting landmarks, cultural relevance, or the scenic beauty of a location. The tone is informative and inviting, meant to engage and educate the audience about the place.}\\ \hline
Psychologist & \texttt{\footnotesize Analyzes the psychological states of people or the emotional tone of the scene. The description is introspective, focusing on body language, facial expressions, and possible underlying emotions or tensions.}\\ \hline
Travel Blogger & \texttt{\footnotesize Describes the experience of being in a place, often using enthusiastic, sensory-rich language. The focus is on personal experience, beauty, and what makes the location worth visiting.}\\ \hline
Mechanic & \texttt{\footnotesize Focuses on the technical and mechanical aspects of machines or vehicles, describing their functions, condition, and efficiency. The language is often practical and concerned with performance and maintenance.}\\ \hline
Biologist & \texttt{\footnotesize Describes natural elements such as flora, fauna, ecosystems, or weather, with attention to scientific classifications and behaviors. The focus is on living organisms and their interaction with the environment.}\\ \hline
Detective & \texttt{\footnotesize Analyzes the scene for clues, providing detailed observations that suggest a narrative or mystery. The language is often investigative, with an emphasis on small, telling details that could reveal a larger story.}\\ \hline
    \end{tabular}
    \caption{Overview of personas and respective prompt description for explicit diversification. Table continues in Tab.~\ref{tab:personas_2}.}
    \label{tab:personas}
\end{table*}
\begin{table*}[t]
 \centering
      \begin{tabular}{p{0.15\linewidth} | p{.85\linewidth}}
     \textbf{Persona}  & \textbf{Description}\\ \hline
 Meteorologist & \texttt{\footnotesize Focuses on weather patterns, climate conditions, and atmospheric details. The description often includes information about wind, temperature, and the impact of weather on the environment.}\\ \hline
 Interior Designer & \texttt{\footnotesize Describes the layout, furniture, color schemes, and design elements of a room or space. The focus is on how the design choices create a functional and aesthetically pleasing environment.}\\ \hline
 Engineer & \texttt{\footnotesize Focuses on the design, functionality, and technical specifications of structures, systems, or machinery. Descriptions emphasize problem-solving, durability, and how elements work together to achieve a purpose.}\\ \hline
    \end{tabular}
    \caption{Continuation of Tab.~\ref{tab:personas}}
    \label{tab:personas_2}
\end{table*}

\section{Bias}\label{app:bias}
\begin{figure*}[t]
     \centering
     \begin{subfigure}[b]{0.48\textwidth}
         \centering
        \includegraphics[width=\textwidth]{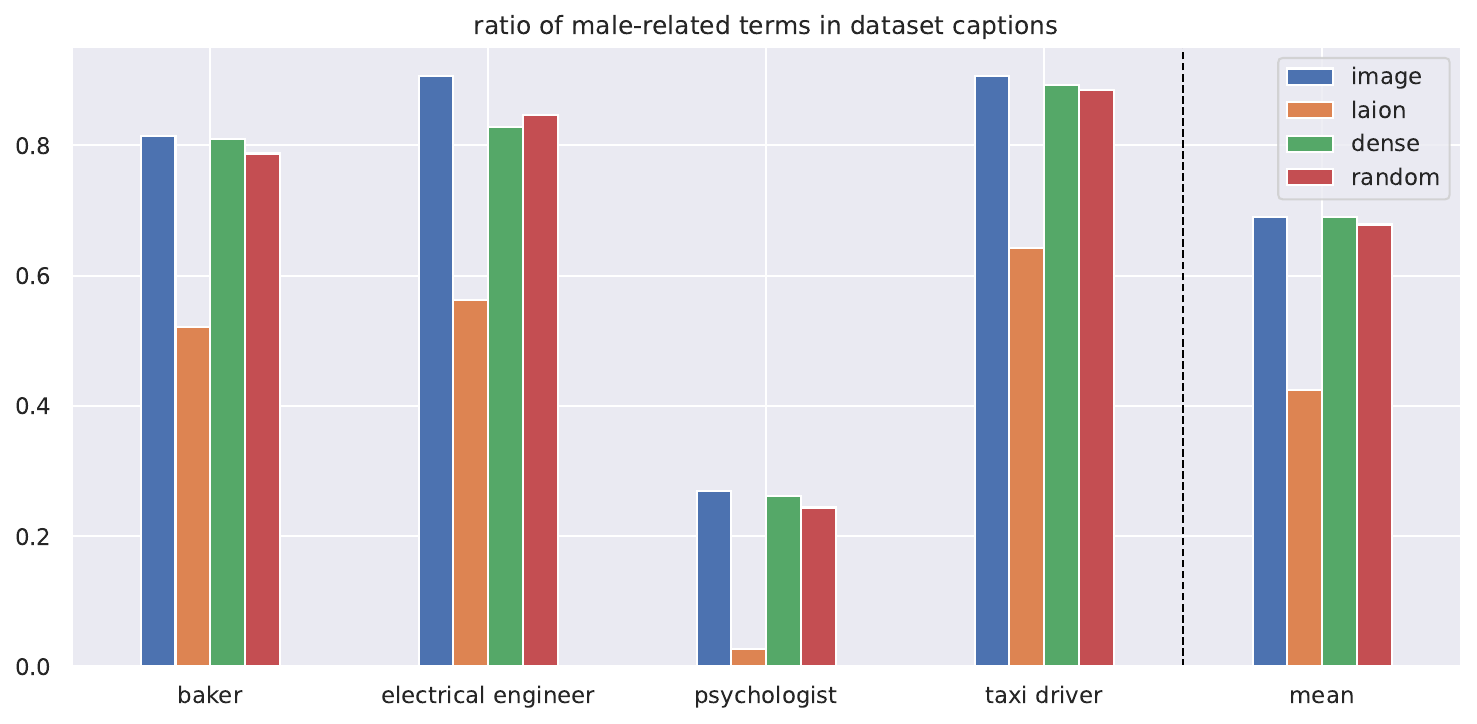}
         \caption{Gender ratio in captions}
         \label{fig:bias_inspection_cap}
     \end{subfigure}
     \hfill
     \begin{subfigure}[b]{0.48\textwidth}
         \centering
        \includegraphics[width=\textwidth]{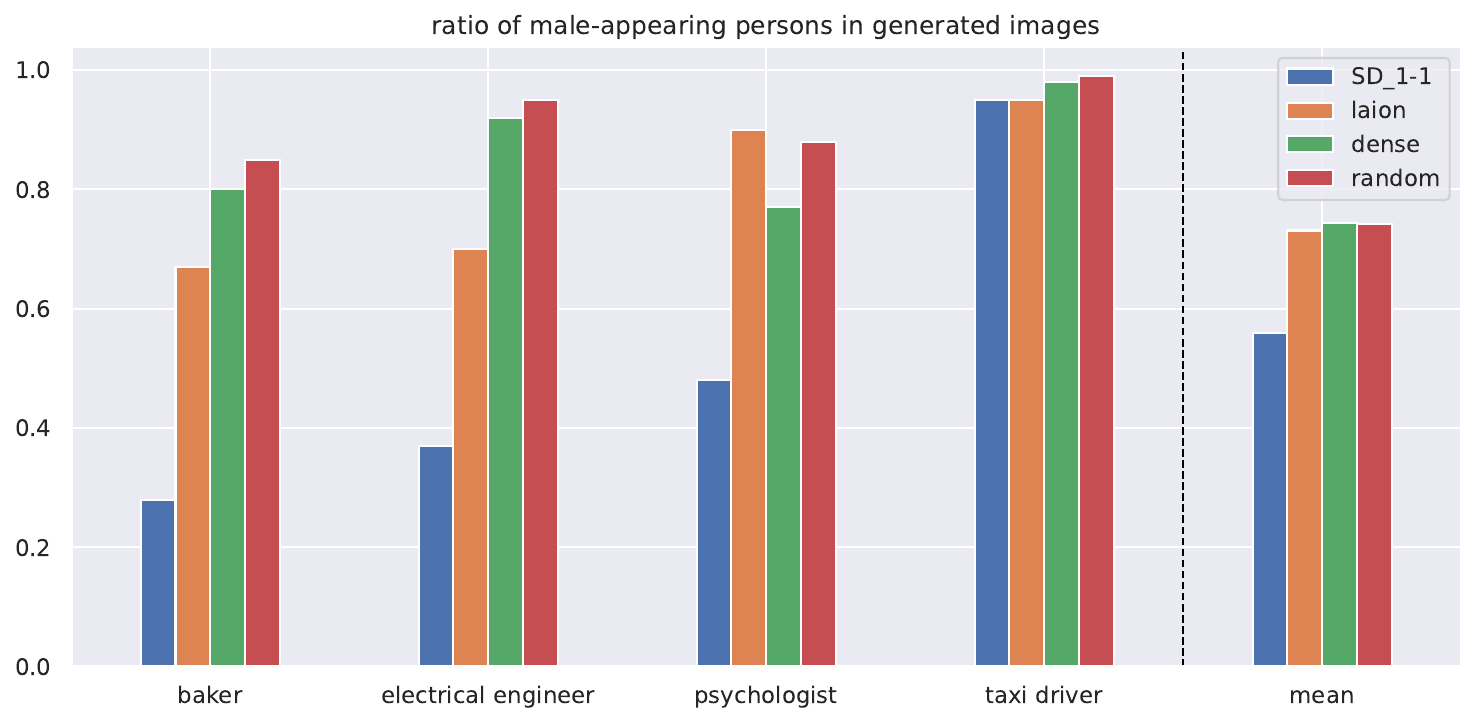}
         \caption{Gender ratio in generated images}
         \label{fig:bias_inspection_imgs}
     \end{subfigure}
     \hfill
    \caption{Bias investigation across four occupations. (a) depicts the training image gender distribution (blue), and the other bars depict the caption distribution. Generally, both synthetic captions (green and red) follow the image distribution well, whereas original Laion captions are much less well-describing and deviate substantially from the image distribution.
    (b) depicts the gender distribution in generated images with baseline SDv1.1 and the tuned versions.
    After continual pre-training (right side, blue vs.~other bars), we observe an increase in male-dominated bias. The generated images follow the training image and caption distribution very well except for ``psychologist'', where there is a complete mismatch between training and outcome distributions.}
    \label{fig:bias_inspection}
\end{figure*}

This section provides additional details on our experiments relating to model bias. In the main body, we showed that bias behavior varies across checkpoints, even if all were trained on the same images. Using different training captions led to significant shifts in the gender ratio of the generated images.

Fig.~\ref{fig:bias_inspection} presents more nuanced results, illustrating four random occupations alongside the average distribution. 
We included these four occupations for clear visualization purposes, but the findings here generalize to the entire set of 150 occupations.

On the left (Fig.~\ref{fig:bias_inspection_cap}), we examine the gender ratio within the training dataset. The blue bars represent the gender ratio in the images, while the orange bars correspond to the ratio in the original LAION captions. The green and red bars show the gender ratios in dense and random-length captions generated with InternVL2, respectively. Consistent with the main results, the gender distributions for dense and random-length captions are closely aligned for individual examples and the average distribution. These distributions are also similar to the image gender ratio (blue), indicating that the captioning model effectively annotated the images and accurately captured respective gender expression. In contrast, the original LAION captions exhibit considerable noise and deviate significantly from the image distribution and synthetic caption distributions. Quantitative results further support these findings: for 91\% of the occupations, the synthetic caption distributions differ no more than 5\% from each other. Conversely, only around 10\% of occupations show a difference of 5\% or less between LAION and synthetic caption distributions.

On the right (Fig.~\ref{fig:bias_inspection_imgs}), we explore the gender ratio of images generated by the models. The blue bars depict results for the baseline SDv1.1 model, while the other bars represent results for SDv1.1 fine-tuned on original LAION captions (orange), dense captions (green), and random-length (red) captions, respectively. Once again, we observe that the models trained on synthetic captions produce similar gender ratios in their outputs, whereas the model trained on LAION captions differs significantly. These results indicate that gender distribution in the caption strongly influences the output bias of a model trained on them.

Interestingly, the caption distribution (left figure) does not clearly predict occupations' output distribution (right figure). Looking at the occupational examples of ``baker'', ``electrical engineer'', and ``taxi driver'' we find that the gender ratio in the generated images roughly follows the image and synthetic caption distribution in the dataset. However, the generated distribution from the LAION model is substantially different from its caption distribution. Furthermore, for ``psychologist'', we find that while the gender ratio in dataset captions and images is below 30\%, in generated images, the ratio is around 80\%. We found that this occupation is underrepresented in the dataset, leading us to assume that the training distribution is less reflective of the outcome distribution.

Upon further analysis, we observe that all three of our checkpoints tend to generate more male-appearing images than baseline SDv1.1. Though unwanted, this seems expected for models trained on synthetic captions (green/red), as those images and captions are already overall more male-related. However, despite LAION captions containing a significantly lower proportion of male-related content, the model trained on them still shows a strong tendency to generate male-appearing individuals.
We hypothesize that the trained model's gender bias is mainly influenced by the respective image distribution, although the correlation is not particularly strong. These findings suggest that the resulting output distribution cannot be easily predicted or controlled by the image or caption distribution alone. Instead, there seems to be a more complex interplay between the images, captions, and the various pre-trained components (such as the text encoder, e.g., CLIP or T5) of the T2I model.

\end{document}